\documentclass[11pt]{article}

\usepackage[final]{acl}

\usepackage{times}
\usepackage{latexsym}

\usepackage[T1]{fontenc}

\usepackage[utf8]{inputenc}

\usepackage{microtype}

\usepackage{inconsolata}

\usepackage{graphicx}
\usepackage{booktabs}
\usepackage{amsmath}
\usepackage{enumitem}

\title{MINCE: Shrinking LLM Evaluation Datasets via Few-Model Monte Carlo Calibration}

\author{
Devleena Das \quad Rajeev Patwari \quad Vikram Kumar Bukka \quad Nithin Kumar Guggilla 
\AND \quad Elliott Delaye  \quad Ashish Sirasao \\ Advanced Micro Devices (AMD)
}

\begin{document}
\maketitle
\begin{abstract}
    Evaluating LLMs across many model variants---quantized, fine-tuned, or deployment-specific---requires running large benchmarks repeatedly, a process that can take tens of hours per model on edge hardware such as NPUs. Existing subset selection methods reduce this cost but depend on large calibration pools or learned prediction layers.
    We introduce MINCE (Monte Carlo Informed N-sizing for Compact Evaluation), which uses Monte Carlo simulation over per-item logs from a small set of calibration models to find the minimum subset size that bounds accuracy drift and then fixes a randomly sampled subset at that size, with no prediction layer needed.
    MINCE reduces IFEVAL by 54\%, MMLU by 89\%, GSM8K by 70\%, and MMLU-Pro by 88\% with maximum drift $\leq$2.62\,pp on BF16 calibration models. The frozen subsets generalize to held-out GPU models with mean drift $\leq$1.40\,pp and to INT4 NPU models with mean drift of 0.77--3.59\,pp, while delivering evaluation speedups of up to 8.1$\times$ on the GPU models and evaluation speedups of 1.7--3.4$\times$ on the NPU models.
    The method is robust to calibration pool size and achieves lower drift than tinyBenchmarks (12$\times$ lower on MMLU, 3.3$\times$ on GSM8K) while using 42$\times$ fewer calibration models.
\end{abstract}

\section{Introduction}
\label{sec:introduction}

Rigorous evaluation of large language models (LLMs) requires running increasingly large benchmarks across a growing number of model variants.
Standard benchmarks range from hundreds of items (IFEVAL~\citep{zhou2023ifeval}, 541; GSM8K~\citep{cobbe2021gsm8k}, 1,319) to tens of thousands (MMLU~\citep{hendrycks2021mmlu}, 14,042), and practitioners must run them across many model variants, such as quantized, fine-tuned, each requiring validation before deployment.
This cost is especially acute on edge accelerators such as NPUs, where inference throughput is orders of magnitude lower than on datacenter GPUs. For example, industry inference benchmarking suites such as MLPerf Client and Mobile~\citep{reddi2020mlperf} utilize multiple benchmarks across many model variants on edge hardware, making full evaluation a major bottleneck for model iteration.

Several methods reduce evaluation cost through benchmark subsetting using IRT, genetic algorithms, sparse optimization, clustering, or surrogate models~\citep{polo2024tinybenchmarks,kipnis2025metabench,wang2026essencebench,zhang2026sparseeval,yuan2025tailoredbench}. Many of these methods require large calibration pools or introduce learned subset selection.
For newer benchmarks, such calibration pools may not exist, and assembling them for each new benchmark is impractical for scalability.
This raises a practical question: \textit{can benchmark subsets be determined reliably from a small number of calibration models, without learned item selection or prediction layers?}

In this work, we propose MINCE (\textbf{M}onte Carlo \textbf{I}nformed \textbf{N}-sizing for \textbf{C}ompact \textbf{E}valuation), which reframes benchmark subsetting as asking \textit{how many} items to select rather than \textit{which} ones. We posit that if enough items are included to keep subset scores close to full-benchmark scores, the specific choice of items matters less. To this end, MINCE determines the sufficient subset size for a given benchmark and fixes a randomly sampled subset at that size, requiring no learned item selector.
Specifically, given evaluation logs from 7 models in our main setting, MINCE uses Monte Carlo simulation~\citep{rubinstein2008simulation} to measure worst-case accuracy drift, the gap between subset and full-benchmark scores, at multiple candidate subset sizes and identifies the point of diminishing returns where adding more items no longer meaningfully reduces drift. The resulting subset size is benchmark-specific.

We evaluate MINCE across four benchmarks (IFEVAL, MMLU, GSM8K, MMLU-Pro) using 7 BF16 calibration models spanning 4 architecture families, and validate that the frozen subsets generalize to 4 held-out GPU models (spanning up to 141B, and with FP8/FP8-KV quantization) and 3 held-out INT4 NPU models.
Our contributions are:
\begin{enumerate}[leftmargin=*,nosep]
    \item We introduce MINCE, requiring only 7 calibration models to reduce IFEVAL/MMLU/GSM8K/MMLU-Pro by 54/89/70/88\% with drift $\leq$2.62\,pp on BF16 models. The frozen subsets generalize to held-out GPU models (mean drift $\leq$1.40\,pp, up to 8.1$\times$ speedup) and INT4 NPU models (mean drift 0.77--3.59\,pp, 1.7--3.4$\times$ speedup) (Section~\ref{sec:overall_results}).
    \item Compared to tinyBenchmarks~\citep{polo2024tinybenchmarks}, MINCE achieves 12$\times$ lower mean drift on MMLU and 3.3$\times$ on GSM8K while using 42$\times$ fewer calibration models (Section~\ref{sec:tinybenchmarks}).
    \item We provide practical calibration guidelines by testing the robustness of MINCE as the calibration pool shrinks. The selected subset size remains stable even when reducing from 7 to 3 calibration models on MMLU/GSM8K, giving practitioners flexibility to scale calibration effort to their evaluation scenario (Section~\ref{sec:ablation_calibration}).
\end{enumerate}

\section{Related Work}
\label{sec:related}

\paragraph{IRT-based subset selection.}
Several methods use Item Response Theory (IRT) to select informative benchmark items.
tinyBenchmarks~\citep{polo2024tinybenchmarks} draws on item-level responses from a pool of 395 previously evaluated models, using 75\% (approximately 296 models) per experimental split to fit IRT models and select 100 anchor items via k-means clustering of IRT-based item embeddings. It then uses an IRT-based estimator to estimate full-benchmark performance from responses on those items. 
MetaBench~\citep{kipnis2025metabench} calibrates IRT on $>$5,000 models, selects items via Fisher information, and reconstructs scores with a Generalized Additive Model surrogate.
In contrast, MINCE uses only 7 calibration models in our main setting and uses Monte Carlo simulation to find the subset size at which raw subset accuracy reliably approximates the full-benchmark score.

\paragraph{Optimization-based approaches.}
Another area of work uses optimization or proxy-based selection methods to determine which items to retain in a subset.
EssenceBench~\citep{wang2026essencebench} uses a genetic algorithm to search over candidate subsets, scoring each by how well a surrogate model reproduces full-benchmark rankings.
SparseEval~\citep{zhang2026sparseeval} formulates subset selection as sparse optimization over the model--item performance matrix using a learned MLP proxy.
TailoredBench~\citep{yuan2025tailoredbench} constructs model-specific coresets via K-Medoids clustering over a large reference pool of pre-evaluated models.
These methods typically require either large pools of pre-evaluated models or learned optimization/proxy models.
Moreover, these methods primarily focus on \emph{which} items to keep under a given evaluation budget, whereas MINCE instead asks \emph{how many} items are needed for accurate estimation of full datasets.
Our ablations suggest that, at the subset sizes selected by MINCE, random sampling performs comparably to stratified and embedding-clustered sampling removing the need for large calibration pools, specialized selectors, or learned score-reconstruction layers (see Section~\ref{sec:ablation_strategy}).

\section{Method}
\label{sec:method}

\begin{figure*}[t]
    \centering
    \includegraphics[width=0.99\textwidth]{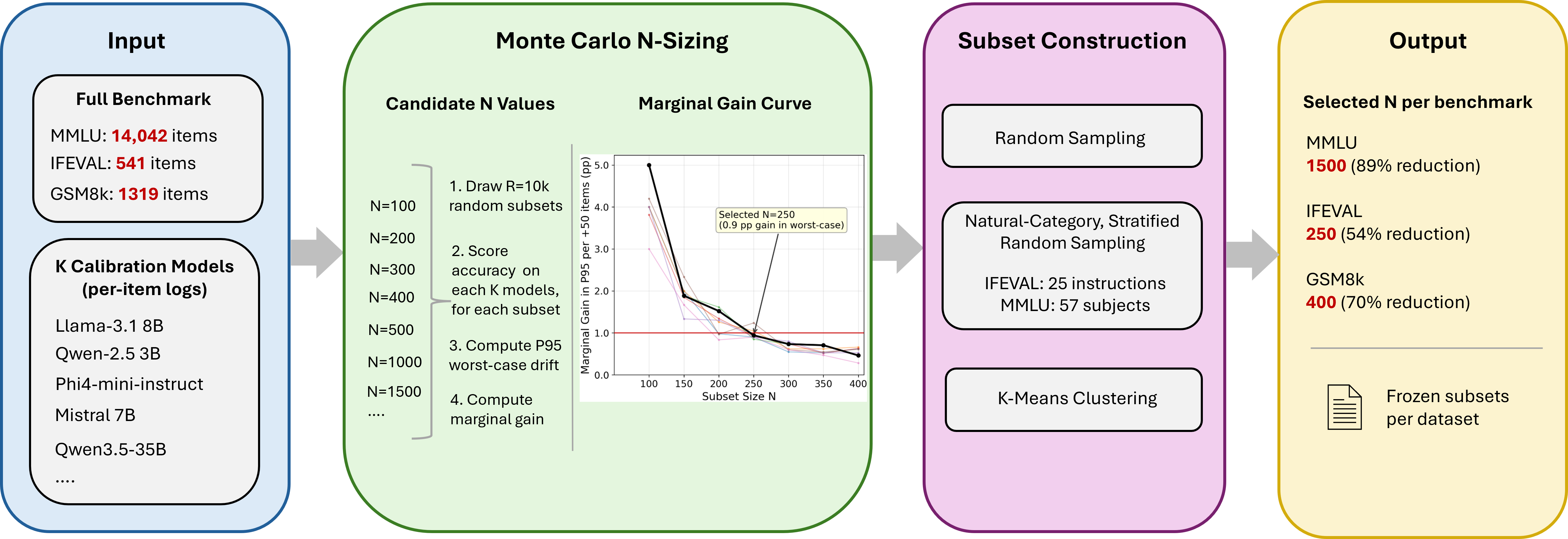}
    \caption{Overview of the MINCE pipeline. Per-item logs from $K$ models are used to identify $n^{*}$ via Monte Carlo simulation. At $n^{*}$, three tested sampling strategies (uniform, stratified, k-means) yield comparable drift, allowing MINCE to use random sampling by default. The final output is a frozen evaluation subset  smaller than the original.}
    \label{fig:pipeline}
\end{figure*}

Figure~\ref{fig:pipeline} shows the MINCE pipeline. MINCE requires, as input, per-item evaluation logs from a small set of calibration models that have already been evaluated on the full benchmark.
The pipeline then proceeds in two stages:
(1)~\textbf{Monte Carlo sizing} determines the minimum subset size $n^{*}$ at which accuracy drift between the subset and the full benchmark is bounded (Section~\ref{sec:mc});
(2)~\textbf{Subset construction} samples $n^{*}$ items and outputs a frozen subset file reused for all future evaluations of that benchmark (Section~\ref{sec:practical}).
\subsection{Problem Formulation}
\label{sec:formulation}

Let $\mathcal{B}$ denote a benchmark with $N$ items and let $\mathcal{M} = \{m_1, \ldots, m_K\}$ be a set of $K$ calibration models. Each model $m_k$ has been evaluated on every item in $\mathcal{B}$, producing per-item pass/fail results. For a subset $S \subseteq \mathcal{B}$ of size $n$, \textit{accuracy drift} is defined as:
\begin{equation}
    \delta(S, m_k) = |\text{acc}(S, m_k) - \text{acc}(\mathcal{B}, m_k)|
\end{equation}
where $\text{acc}(\cdot, m_k)$ denotes model $m_k$'s accuracy on the given item set. For benchmarks with multiple metrics, such as IFEVAL~\citep{zhou2023ifeval}, drift is computed per metric and the worst case is taken.
The \textit{worst-case drift} across all calibration models is defined as:
\begin{equation}
    \delta^{*}(S) = \max_{k} \, \delta(S, m_k)
    \label{eq:worst_drift}
\end{equation}

The goal of MINCE is to find the smallest subset size $n^{*}$ at which adding more items no longer meaningfully reduces worst-case drift.
Concretely, we sweep candidate sizes, measure drift at each candidate subset size, and select $n^{*}$ as the first size where the \textit{marginal gain} drops below a threshold $\tau$:
\begin{equation}
    n^{*} = \min_{t} \; n_t \;\; \text{s.t.} \;\; \Delta(n_t) < \tau .
    \label{eq:nstar}
\end{equation}
where $\Delta(n_t) = \text{P95}(n_{t-1}) - \text{P95}(n_t)$ is the \textit{marginal gain}, i.e., the reduction in 95th-percentile drift between consecutive sizes.
Classical sample size formulas~\citep{cochran1977sampling} assume infinite populations, independent draws, and a single metric, which do not hold for many LLM benchmarks. We therefore use Monte Carlo simulation to determine $n^{*}$ empirically (Section~\ref{sec:mc}).

\subsection{Monte Carlo Sample Size Estimation}
\label{sec:mc}

As shown in the first stage of Figure~\ref{fig:pipeline}, the sizing procedure sweeps candidate sizes $\{n_1, \ldots, n_T\}$. For each $n_t$, we draw $R{=}$10,000 random subsets uniformly without replacement. For each subset, we compute $\delta^{*}(S)$ (Eq.~\ref{eq:worst_drift}) and retain the worst-case drift across all calibration models. Taking the 95th percentile of these worst-case drifts across all $R$ draws gives $\text{P95}(n_t)$, meaning 95\% of random subsets at size $n_t$ produce worst-case drift below $\text{P95}(n_t)$.
We sweep candidate sizes at fixed increments of 50 for IFEVAL, 100 for GSM8K, and 500 for MMLU.
We then select $n^{*}$ as the first size where the marginal gain drops below $\tau$ (Eq.~\ref{eq:nstar}).
In practice, we set $\tau{=}$1\,pp per step. We show in Section~\ref{sec:n_sizing} that marginal gain diminishes beyond this threshold, making 1\,pp a practical choice. Appendix~\ref{app:tau_sensitivity} provides a sensitivity analysis confirming that $n^{*}$ remains stable at neighboring $\tau$ values.
The step size between consecutive candidate sizes is a practitioner choice that controls the resolution of this sweep, analogous to a grid spacing in a hyperparameter search. The resulting $n^{*}$ is validated empirically in Section~\ref{sec:results}.

\subsection{Subset Construction}
\label{sec:practical}

Once $n^{*}$ is determined, items are sampled via uniform random sampling, stratified random sampling across natural categories~\citep{balinski2010fair}, or k-means clustering on sentence embeddings.
As we show in Section~\ref{sec:ablation_strategy}, all three produce comparable drift at $n^{*}$, because the Monte Carlo sizing already ensures bounded variance for random draws.
This is a key strength of MINCE, that once the right subset size is determined, random sampling suffices and stratified sampling or learned sampling like k-means are not required.

\section{Experimental Setup}
\label{sec:setup}

We validate MINCE on benchmarks spanning multiple-choice, instruction-following, and chain-of-thought math, with evaluations conducted on both GPUs (AMD Instinct MI210 and MI300X) and edge NPUs (AMD RyzenAI).

\paragraph{Benchmarks.}
\textbf{IFEVAL}~\citep{zhou2023ifeval} (541 items, generative instruction-following),
\textbf{MMLU}~\citep{hendrycks2021mmlu} (14,042 items, multiple-choice, 5-shot),
\textbf{GSM8K}~\citep{cobbe2021gsm8k} (1,319 items, 0-shot).
For the out-of-sample GPU evaluation (Section~\ref{sec:overall_results}), we additionally evaluate on \textbf{MMLU-Pro}~\citep{wang2024mmlu} (12,032 items, 5-shot), a harder multiple-choice benchmark where MMLU saturates for larger models.
All benchmark accuracies are collected using lm-eval-harness~\citep{eval-harness}.
In practice, we fix seed\,42 for subset construction; Section~\ref{sec:ablation_strategy} shows that drift remains within bounds across 100 random seeds.

\paragraph{Calibration models.}
We leverage 7 BF16 models spanning 3B--35B parameters and 4 architecture families:
Llama-3.1-8B (base) and Instruct~\citep{grattafiori2024llama3},
Qwen2.5-3B-Instruct, Qwen3-8B, and Qwen3.5-35B-A3B (MoE)~\citep{yang2025qwen3},
Mistral-7B-Instruct~\citep{jiang2023mistral},
and Phi-4-mini-Instruct~\citep{abdin2024phi4}.
These models span 25\%--89\% accuracy across the benchmarks, ensuring the calibration pool covers both weak and strong model performers. In Section~\ref{sec:ablation_calibration}, we further experiment with whether less than 7 models can suffice for stable $n^{*}$ estimation.

\paragraph{Validation.}
We validate on three model groups:
(1)~the 7 BF16 calibration models, verifying drift at $n^{*}$ stays within an expected bound;
(2)~four out-of-sample GPU models spanning larger sizes and FP8 quantization: Hermes-4-70B and its FP8 variant, Llama-3.1-8B-Instruct-FP8-KV, and Mixtral-8x22B-Instruct-FP8-KV (141B MoE);
(3)~three held-out INT4 NPU variants (Llama-3.1-8B-Inst, Phi-4-mini, Qwen3-8B), testing generalization to unseen quantization and hardware.

\paragraph{K-means clustering.}
For benchmarks without natural categories, and to test whether a learned sampling strategy offers an advantage, we evaluate k-means sampling via clustering on sentence embeddings from Qwen3-Embedding-8B~\citep{zhang2025qwen3embedding} (details in Appendix~\ref{app:kmeans}).

\section{Results}
\label{sec:results}
\subsection{MINCE Overall Results}
\label{sec:overall_results}

Table~\ref{tab:table1} presents full and $n^{*}$ subset accuracy for all 7 BF16 models, along with per-model accuracy drift ($|\delta|$) and aggregate mean and max drift across all models.
Overall, MINCE reduces IFEVAL by 54\%, MMLU by 89\%, GSM8K by 70\%, and MMLU-Pro by 88\%, while bounding accuracy drift to a mean of 0.44--1.40\,pp and a max of $\leq$2.62\,pp across all models. These varying reduction ratios show that MINCE is benchmark-specific and does not assume every benchmark can be reduced by the same amount. Instead, the Monte Carlo sizing determines $n^{*}$ independently for each benchmark based on its observed drift characteristics.
Model rankings are also preserved: Spearman $\rho = 1.0$ on IFEVAL and GSM8K, and $0.96$ on MMLU and MMLU-Pro.
Note, table~\ref{tab:table1} reports IFEVAL drift on instruction-strict as a summary. Full per-metric breakdowns for all 4 IFEVAL metrics are in Appendix~\ref{app:permodel} with no change in conclusions.
Tables~\ref{tab:npu} and~\ref{tab:oos_gpu} show that these results transfer beyond the BF16 calibration models and that the frozen $n^{*}$ subsets generalize to held-out INT4 NPU models and to larger GPU models under FP8 quantization.

\begin{table*}[t]
\centering
\footnotesize
\setlength{\tabcolsep}{2pt}
\resizebox{\textwidth}{!}{%
\begin{tabular}{l ccc ccc ccc ccc}
\toprule
& \multicolumn{3}{c}{\textbf{IFEVAL} ($n^{*}{=}250$, 54\%$\downarrow$)} & \multicolumn{3}{c}{\textbf{MMLU} ($n^{*}{=}1500$, 89\%$\downarrow$)} & \multicolumn{3}{c}{\textbf{GSM8K} ($n^{*}{=}400$, 70\%$\downarrow$)} & \multicolumn{3}{c}{\textbf{MMLU-Pro} ($n^{*}{=}1500$, 88\%$\downarrow$)} \\
\cmidrule(lr){2-4} \cmidrule(lr){5-7} \cmidrule(lr){8-10} \cmidrule(lr){11-13}
\textbf{Model} & Full & $n^{*}$ & $|\delta|$ & Full & $n^{*}$ & $|\delta|$ & Full & $n^{*}$ & $|\delta|$ & Full & $n^{*}$ & $|\delta|$ \\
\midrule
Llama-3.1-8B (base) & 35.13 & 34.03 & 1.10 & 65.39 & 65.00 & 0.39 & 25.63 & 25.00 & 0.63 & 35.63 & 35.07 & 0.56 \\
Qwen2.5-3B-Inst & 69.30 & 70.16 & 0.85 & 65.27 & 66.47 & 1.20 & 45.56 & 44.00 & 1.56 & 44.23 & 43.53 & 0.70 \\
Mistral-7B-Inst & 57.91 & 57.07 & 0.85 & 60.87 & 61.73 & 0.86 & 43.37 & 40.75 & 2.62 & 35.55 & 35.40 & 0.15 \\
Phi-4-mini-Inst & 75.66 & 77.49 & 1.83 & 67.90 & 68.27 & 0.37 & 75.66 & 74.50 & 1.16 & 46.56 & 45.07 & 1.49 \\
Llama-3.1-8B-Inst & 80.94 & 80.63 & 0.31 & 68.77 & 68.87 & 0.09 & 73.54 & 71.00 & 2.54 & 46.91 & 46.73 & 0.17 \\
Qwen3-8B & 88.85 & 87.43 & 1.41 & 72.01 & 71.93 & 0.08 & 56.10 & 55.25 & 0.85 & 63.17 & 63.13 & 0.04 \\
Qwen3.5-35B-A3B & 88.97 & 89.53 & 0.56 & 83.29 & 83.20 & 0.09 & 54.44 & 54.00 & 0.44 & 75.90 & 77.20 & 1.30 \\
\midrule
Mean $|\delta|$ (pp) & \multicolumn{3}{c}{0.99} & \multicolumn{3}{c}{0.44} & \multicolumn{3}{c}{1.40} & \multicolumn{3}{c}{0.63} \\
Max $|\delta|$ (pp) & \multicolumn{3}{c}{1.83} & \multicolumn{3}{c}{1.20} & \multicolumn{3}{c}{2.62} & \multicolumn{3}{c}{1.49} \\
\bottomrule
\end{tabular}%
}
\caption{Full-benchmark, $n^{*}$-subset accuracy (\%) and absolute drift $|\delta|$ for 7 BF16 calibration models across 4 benchmarks.}
\label{tab:table1}
\end{table*}

\begin{table*}[t!]
\centering
\footnotesize
\setlength{\tabcolsep}{3pt}
\begin{tabular}{l ccc ccc ccc cc}
\toprule
& \multicolumn{3}{c}{\textbf{IFEVAL}} & \multicolumn{3}{c}{\textbf{MMLU}} & \multicolumn{3}{c}{\textbf{GSM8K}} & \multicolumn{2}{c}{\textbf{Speedup ($\times$)}} \\
\cmidrule(lr){2-4} \cmidrule(lr){5-7} \cmidrule(lr){8-10} \cmidrule(lr){11-12}
\textbf{NPU Model (INT4)} & Full & $n^{*}$ & $|\delta|$ & Full & $n^{*}$ & $|\delta|$ & Full & $n^{*}$ & $|\delta|$ & IFEVAL & MMLU \\
\midrule
Llama-3.1-8B-Inst & 26.74 & 29.06 & 2.32 & 62.65 & 64.13 & 1.48 & 15.23 & 16.25 & 1.02 & 1.6$\times$ & 3.4$\times$ \\
Phi-4-mini-Inst & 62.35 & 54.18 & 8.17 & 58.95 & 59.67 & 0.72 & 67.32 & 66.50 & 0.82 & 2.1$\times$ & 2.0$\times$ \\
Qwen3-8B & 84.53 & 84.82 & 0.29 & 73.57 & 73.47 & 0.10 & 58.98 & 58.50 & 0.48 & 1.7$\times$ & 1.7$\times$ \\
\midrule
Mean $|\delta|$ (pp) & & & 3.59 & & & 0.77 & & & 0.77 & 1.8$\times$ & 2.4$\times$ \\
Max $|\delta|$ (pp) & & & 8.17 & & & 1.48 & & & 1.02 & & \\
\bottomrule
\end{tabular}
\caption{Out-of-sample validation on 3 INT4 NPU models held out from calibration. Full and $n^{*}$-subset accuracy (\%), absolute drift $|\delta|$ (pp), and evaluation speedup on AMD Ryzen AI.}
\label{tab:npu}
\end{table*}

\paragraph{Out-of-Sample NPU Validation} Table~\ref{tab:npu} demonstrates that BF16-calibrated subsets generalize across the quantization gap by measuring drift on 3 INT4 NPU models held out from all calibration. 
Mean $|\delta|$ is 3.59\,pp on IFEVAL (inst\_strict), 0.77\,pp on MMLU, and 0.77\,pp on GSM8K. MMLU and GSM8K drift remains low across all models ($\leq$1.48\,pp); IFEVAL drift is higher for Phi-4-mini-Inst (8.17\,pp). Notably, the same model drifts only 0.72\,pp on MMLU and 0.82\,pp on GSM8K, indicating that the outlier is specific to the combination of INT4 quantization effects and IFEVAL's generative instruction-following task. For deployment-critical use cases where such task-specific sensitivity may arise, practitioners should consider tightening the marginal-gain threshold, applying per-metric calibration, or retaining a larger subset.

\paragraph{Out-of-Sample GPU Validation} Table~\ref{tab:oos_gpu} demonstrates that BF16-calibrated subsets generalize to larger and reduced-precision GPU models held out from calibration: Hermes-4-70B and its FP8 variant, Llama-3.1-8B-Instruct-FP8-KV, and Mixtral-8x22B-Instruct-FP8-KV (141B MoE), quantized via AMD Quark\footnote{\url{https://github.com/amd/quark}}.
Mean $|\delta|$ is 0.70\,pp on MMLU, 1.55\,pp on GSM8K, and 0.81\,pp on MMLU-Pro, on par with in-sample drift. FP8 quantization does not degrade subset fidelity as Hermes-4-70B-FP8 drifts comparably to its BF16 counterpart.

\begin{table*}[t!]
\centering
\footnotesize
\setlength{\tabcolsep}{2pt}
\resizebox{\textwidth}{!}{%
\begin{tabular}{l ccc ccc ccc ccc}
\toprule
& \multicolumn{3}{c}{\textbf{MMLU} ($n^{*}{=}1500$, 89\%$\downarrow$)} & \multicolumn{3}{c}{\textbf{GSM8K} ($n^{*}{=}400$, 70\%$\downarrow$)} & \multicolumn{3}{c}{\textbf{MMLU-Pro} ($n^{*}{=}1500$, 88\%$\downarrow$)} & \multicolumn{3}{c}{\textbf{Speedup ($\times$)}} \\
\cmidrule(lr){2-4} \cmidrule(lr){5-7} \cmidrule(lr){8-10} \cmidrule(lr){11-13}
\textbf{Model} & Full & $n^{*}$ & $|\delta|$ & Full & $n^{*}$ & $|\delta|$ & Full & $n^{*}$ & $|\delta|$ & MMLU & GSM8K & MMLU-Pro \\
\midrule
Hermes-4-70B & 79.66 & 79.87 & 0.21 & 65.43 & 67.75 & 2.32 & 62.97 & 62.80 & 0.17 & 8.1$\times$ & 1.2$\times$ & 4.8$\times$ \\
Hermes-4-70B-FP8 & 79.13 & 79.47 & 0.33 & 65.28 & 67.00 & 1.72 & 62.77 & 61.53 & 1.23 & 7.6$\times$ & 1.2$\times$ & 3.7$\times$ \\
Llama-3.1-8B-Inst-FP8-KV & 68.47 & 67.07 & 1.41 & 73.01 & 72.50 & 0.51 & 46.83 & 47.60 & 0.77 & 5.1$\times$ & 1.1$\times$ & 3.3$\times$ \\
Mixtral-8x22B-FP8-KV (141B) & 75.46 & 76.33 & 0.87 & 68.61 & 70.25 & 1.64 & 55.40 & 54.33 & 1.07 & 7.2$\times$ & 1.2$\times$ & 3.6$\times$ \\
\midrule
Mean $|\delta|$ (pp) & \multicolumn{3}{c}{0.70} & \multicolumn{3}{c}{1.55} & \multicolumn{3}{c}{0.81} & 7.0$\times$ & 1.2$\times$ & 3.9$\times$ \\
Max $|\delta|$ (pp) & \multicolumn{3}{c}{1.41} & \multicolumn{3}{c}{2.32} & \multicolumn{3}{c}{1.23} & & & \\
\bottomrule
\end{tabular}%
}
\caption{Out-of-sample GPU validation on 4 larger and reduced-precision models not used in calibration. The frozen $n^{*}$ subset is reused without re-sizing. Full and $n^{*}$-subset accuracy (\%), absolute drift $|\delta|$ (pp), and evaluation speedup on AMD MI300X.}
\label{tab:oos_gpu}
\end{table*}

\paragraph{Wall-clock Savings} The benchmark reductions from MINCE yield meaningful evaluation speedups under batch-size-1 sequential inference.
On GPU, median in-sample speedups range from 2.7$\times$ (IFEVAL) to 8.1$\times$ (MMLU), with GSM8K at 3.3$\times$ (see Table~\ref{tab:timing} in the Appendix for per-model breakdowns). Out-of-sample GPU speedups are comparable: MMLU 5.1--8.1$\times$, MMLU-Pro 3.3--4.8$\times$, and GSM8K ${\sim}$1.2$\times$ (Table~\ref{tab:oos_gpu}).
On NPU hardware, median speedups are 1.7$\times$ on IFEVAL and 2.0$\times$ on MMLU, reaching 3.4$\times$ for Llama-3.1-8B-Inst (Table~\ref{tab:npu}).

\subsection{Monte Carlo N-Sizing}
\label{sec:n_sizing}

\begin{figure*}[t!]
    \centering
    \includegraphics[width=0.99\textwidth]{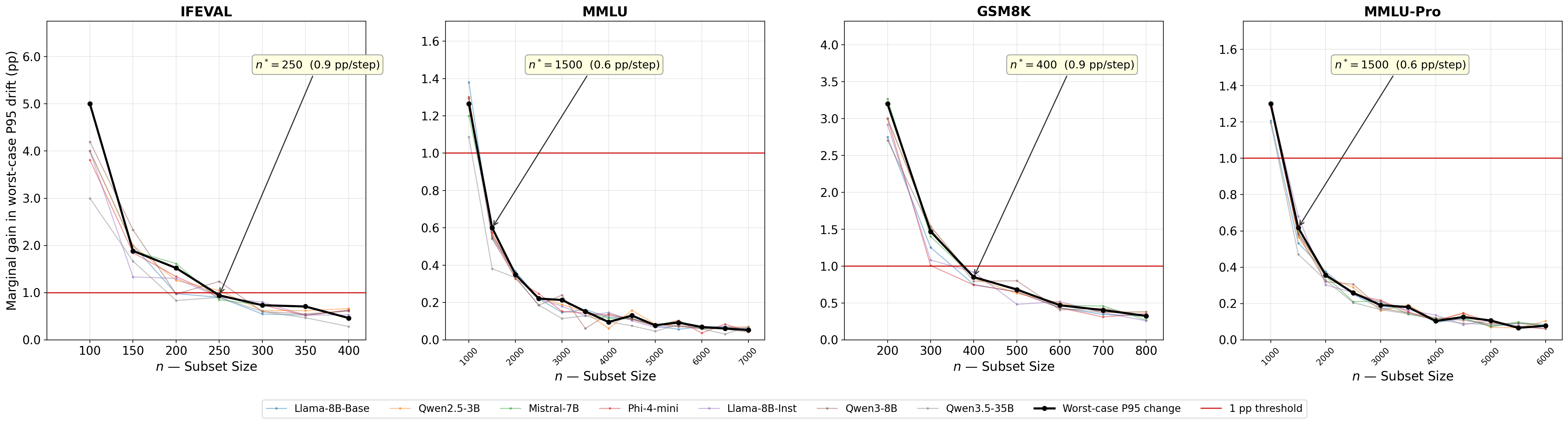}
    \caption{Marginal gain in P95 drift reduction at each candidate subset size. Each curve represents one model; the black curve traces worst case drift. $n^{*}$ is the first size where  worst-case gain drops below $\tau{=}1$\,pp.}
    \label{fig:marginal}
\end{figure*}
Figure~\ref{fig:marginal} shows the marginal gain in P95 drift reduction, achieved by adding additional items at each candidate subset size $n$.
Each curve represents the marginal gain for a single model. The black curve traces the drop in worst-case P95 drift between consecutive subset sizes.
As shown, beyond 1\,pp the marginal gain diminishes sharply with each additional step.
We therefore set $\tau{=}1$\,pp, and $n^{*}$ is the first size where the marginal gain drops below $\tau$.
IFEVAL reaches $n^{*}{=}250$ (marginal gain 0.9\,pp/step), GSM8K reaches $n^{*}{=}400$ (0.9\,pp/step), MMLU converges at $n^{*}{=}1{,}500$ (0.60\,pp/step), and MMLU-Pro at $n^{*}{=}1{,}500$ (0.60\,pp/step).
IFEVAL's $n^{*}$ is driven by prompt\_strict, the highest-variance metric. See Table~\ref{tab:ifeval_allmetrics} for per-metric drift details.

\subsection{Subset Selection Strategy Ablation}
\label{sec:ablation_strategy}

Significant prior work relies on curated, learned, or model-informed item selection~\citep{polo2024tinybenchmarks,perlitz2024efficient,yuan2025tailoredbench}, raising the question of whether sampling strategy still matters once $n^{*}$ has been determined.
To test this, we compare three strategies, uniform random sampling, stratified random sampling across natural categories (IFEVAL: 25 instruction types; MMLU: 57 subjects), and sampling with k-means clusters.
For each strategy, we drew 100 subsets at $n^{*}$ (seeds 0--99) and measured drift on all 7 BF16 calibration models.

\begin{figure*}[t]
    \centering
    \includegraphics[width=0.99\textwidth]{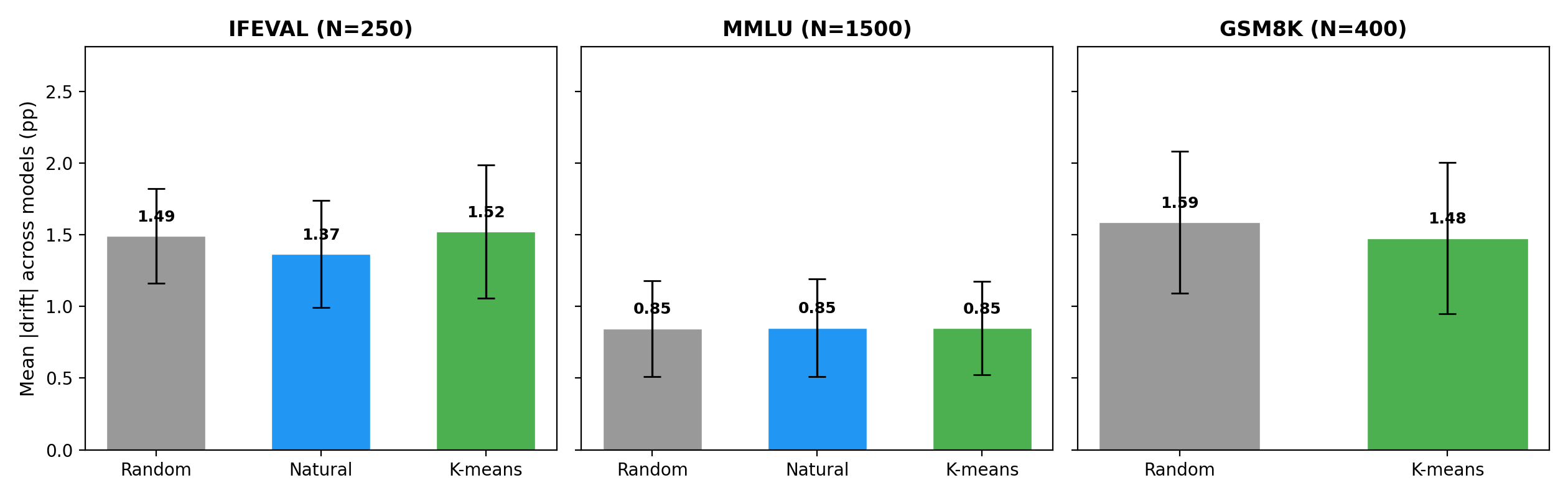}
    \caption{Accuracy drift distributions for uniform, stratified, and k-means sampling strategies at $n^{*}$, evaluated across all 7 BF16 calibration models (Section~\ref{sec:setup}) and over 100 random seeds. All three strategies produce comparable drift, with mean differences below 0.2\,pp.}
    \label{fig:strategy}
\end{figure*}

\begin{figure*}[h]
    \centering
    \includegraphics[width=0.99\textwidth]{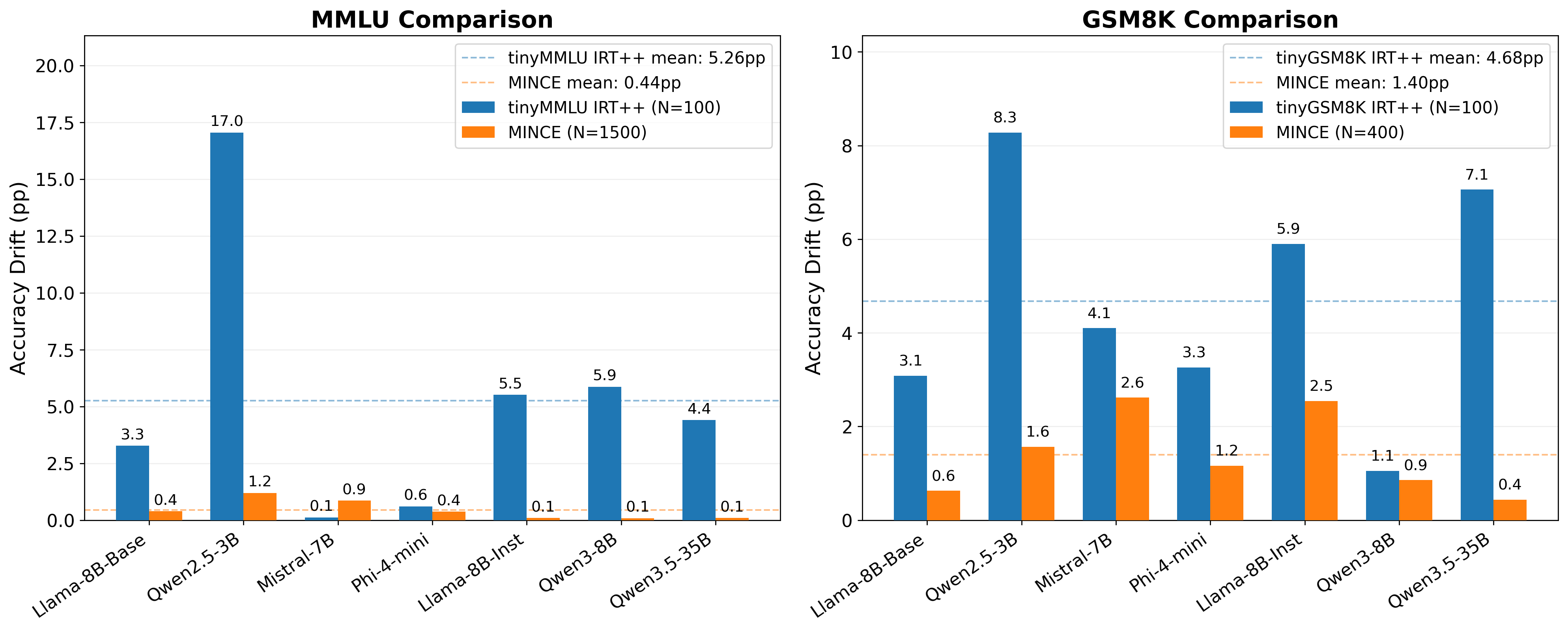}
    \caption{Per-model accuracy drift comparison between tinyBenchmarks and MINCE, showing that MINCE reduces mean drift 12$\times$ on MMLU and 3.3$\times$ on GSM8K while using 42$\times$ fewer calibration models.}
    \label{fig:tinybenchmarks}
\end{figure*}

Figure~\ref{fig:strategy} shows that all strategies produce comparable drift at $n^{*}$, with mean $|\delta|$ ranging from 1.37--1.52\,pp (IFEVAL), 0.85\,pp (MMLU), and 1.48--1.59\,pp (GSM8K).
This result is consistent with the Monte Carlo sizing: at $n^{*}$, random sampling variance is already small enough to bound drift, so stratification has little additional variance to reduce.

\subsection{Calibration Guidelines}
\label{sec:ablation_calibration}

In practice, calibration pool design involves choosing both how many models to include and which architectures to cover. Practitioners evaluating variants of a single model family (e.g., quantized iterations of one architecture) may need only a few calibration models, while those benchmarking across diverse architectures benefit from broader coverage.
To study how MINCE behaves as the calibration pool shrinks, we run a leave-$K$-out analysis. Specifically, we remove $K$ of 7 calibration models and re-run MINCE with the remaining calibration pool for all $\binom{7}{K}$ combinations with $K \in \{1, 2, 3, 4\}$.

\begin{table}[t]
\centering
\footnotesize
\setlength{\tabcolsep}{3pt}
\begin{tabular}{l c c c c}
\toprule
 & $n^{*}$ & $n^{*}$ & Mean & Max \\
\textbf{Benchmark} & \textbf{(7 models)} & \textbf{(3 models)} & $|\delta|$ & $|\delta|$ \\
\midrule
IFEVAL & 250 & 200--300 & 1.13 & 5.94 \\
MMLU & 1500 & 1500 & 0.44 & 1.20 \\
GSM8K & 400 & 400 & 1.40 & 2.62 \\
\bottomrule
\end{tabular}
\caption{Effect of reducing calibration from 7 to 3 models on $n^{*}$, tested across all leave-$K$-out combinations.}
\label{tab:leave_k_out}
\end{table}

Table~\ref{tab:leave_k_out} shows that MMLU and GSM8K are completely stable and $n^{*}$ does not change across any combination. IFEVAL is mostly stable (87/98 combinations) and unstable cases shift $n^{*}$ by $\pm$50 items (see Appendix~\ref{app:leave_k_out}).
This suggests that calibration stability is high, which is a practical advantage of MINCE; practitioners can adjust both the number and composition of calibration models to match evaluation goals without any retraining.

\subsection{Comparison to TinyBenchmarks}
\label{sec:tinybenchmarks}

Among existing compact subsets, tinyBenchmarks~\citep{polo2024tinybenchmarks} is the most widely adopted, integrated into both the Open LLM Leaderboard~\citep{open-llm-leaderboard} and lm-eval-harness. 
We compare with tinyBenchmarks on MMLU and GSM8K via lm-eval-harness. IFEVAL is not supported by tinyBenchmarks and extending it would require assembling a large calibration pool and fitting a benchmark-specific IRT model.
Figure~\ref{fig:tinybenchmarks} shows MINCE achieves 12$\times$ lower mean drift on MMLU (0.44 vs.\ 5.26\,pp) and 3.3$\times$ on GSM8K (1.40 vs.\ 4.68\,pp), with substantially lower max drift (1.20 vs.\ 17.05\,pp; 2.62 vs.\ 8.28\,pp), using 42$\times$ fewer calibration models.
Notably, tinyBenchmarks drift varies widely across models (0.1--17.0\,pp on MMLU), suggesting that the IRT++ prediction layer may still be sensitive to individual model characteristics despite a large calibration pool.
The two methods occupy different points along three axes, including \textit{subset size}, \textit{calibration effort}, and \textit{score reconstruction}. tinyBenchmarks minimizes subset size (100 items) but requires a large calibration pool ($\sim$296 models) and an IRT++ prediction layer to reconstruct full-benchmark scores. MINCE accepts a larger subset but requires only 7 calibration models, no score reconstruction layer, and can be applied to any benchmark, including new or proprietary ones, without pre-existing large-scale calibration data. In practice, tinyBenchmarks is valuable when existing subsets are available and the large calibration pool is already pre-defined, while MINCE is useful for practitioners who want to create compact subsets for new or internal benchmarks using only a small calibration pool.

\section{Conclusion}
\label{sec:conclusion}
As practitioners iterate over many model variants, the time cost of full benchmark evaluation compounds, especially on edge hardware where each run can take tens of hours.
We introduce MINCE, a Monte Carlo approach to efficient benchmarking. Monte Carlo sampling enables MINCE to choose subset sizes at which accuracy drift is bounded, eliminating the need for learned item selection or large calibration pools and reducing benchmark sizes by 54--89\% with drift $\leq$2.62\,pp on BF16 models. The frozen subsets generalize beyond the calibration pool to held-out GPU models and INT4 NPU models without re-sizing, delivering GPU evaluation speedups of up to 8.1$\times$ and NPU speedups of 1.7--3.4$\times$.
Our ablations show that the method is stable across sampling strategies, calibration pool compositions, and threshold choices. Without requiring any learned model or optimization procedure, MINCE gives practitioners direct control over calibration effort, depending on whether singe model evaluations or benchmarking across diverse architectures are needed. As evaluation demands grow with new benchmarks and hardware targets, MINCE offers a scalable foundation for compact evaluation.

\section*{Limitations}
Our calibration pool spans 7 models (3B--35B, 4 families) and does not include frontier-scale (100B+) or non-transformer architectures; therefore, validating MINCE at that scale remains future work.
The 1\,pp marginal gain threshold is a practical heuristic but different applications may warrant stricter or looser values, and sensitivity analysis is important for safety-critical settings.
Each new benchmark requires its own calibration run with prior per-item evaluation logs; for benchmarks whose scoring requires human annotation or expensive external evaluation, this calibration cost remains meaningful.
Also, MINCE produces a frozen subset that is applied uniformly to all models at evaluation time and cannot currently exploit model-specific signal to further reduce items. Exploring hybrid approaches that combine a fixed core with adaptive selection at evaluation time is a promising future direction.


\bibliography{custom}

\begin{thebibliography}{21}
\providecommand{\natexlab}[1]{#1}

\bibitem[{Abdin et~al.(2024)Abdin, Aneja, Awadalla et~al.}]{abdin2024phi4}
Marah Abdin, Jyoti Aneja, Hany Awadalla, and 1 others. 2024.
\newblock Phi-4 technical report.
\newblock \emph{arXiv preprint arXiv:2412.08905}.

\bibitem[{Balinski and Young(2010)}]{balinski2010fair}
Michel~L. Balinski and H.~Peyton Young. 2010.
\newblock \emph{Fair Representation: Meeting the Ideal of One Man, One Vote}, 2nd edition.
\newblock Brookings Institution Press.

\bibitem[{Biderman et~al.(2024)Biderman, Schoelkopf, Anthony, Bradley, O'Brien et~al.}]{eval-harness}
Stella Biderman, Hailey Schoelkopf, Quentin~Gregory Anthony, Herbie Bradley, Kyle O'Brien, and 1 others. 2024.
\newblock \href {https://doi.org/10.5281/zenodo.10256836} {A framework for few-shot language model evaluation}.

\bibitem[{Cobbe et~al.(2021)Cobbe, Kosaraju, Bavarian, Chen, Jun, Kaiser, Plappert, Tworek, Hilton, Nakano, Hesse, and Schulman}]{cobbe2021gsm8k}
Karl Cobbe, Vineet Kosaraju, Mohammad Bavarian, Mark Chen, Heewoo Jun, Lukasz Kaiser, Matthias Plappert, Jerry Tworek, Jacob Hilton, Reiichiro Nakano, Christopher Hesse, and John Schulman. 2021.
\newblock Training verifiers to solve math word problems.
\newblock \emph{arXiv preprint arXiv:2110.14168}.

\bibitem[{Cochran(1977)}]{cochran1977sampling}
William~G. Cochran. 1977.
\newblock \emph{Sampling Techniques}, 3rd edition.
\newblock John Wiley \& Sons.

\bibitem[{Fourrier et~al.(2024)Fourrier, Habib, Wolf, and Tunstall}]{open-llm-leaderboard}
Cl\'{e}mentine Fourrier, Nathan Habib, Thomas Wolf, and Lewis Tunstall. 2024.
\newblock Open {LLM} leaderboard.
\newblock \emph{Hugging Face}.

\bibitem[{Grattafiori et~al.(2024)Grattafiori, Dubey, Jauhri et~al.}]{grattafiori2024llama3}
Aaron Grattafiori, Abhimanyu Dubey, Abhinav Jauhri, and 1 others. 2024.
\newblock The {L}lama 3 herd of models.
\newblock \emph{arXiv preprint arXiv:2407.21783}.

\bibitem[{Hendrycks et~al.(2021)Hendrycks, Burns, Basart, Zou, Mazeika, Song, and Steinhardt}]{hendrycks2021mmlu}
Dan Hendrycks, Collin Burns, Steven Basart, Andy Zou, Mantas Mazeika, Dawn Song, and Jacob Steinhardt. 2021.
\newblock Measuring massive multitask language understanding.
\newblock In \emph{Proceedings of the International Conference on Learning Representations}.

\bibitem[{Jiang et~al.(2023)Jiang, Sablayrolles, Mensch, Bamford, Chaplot, de~las Casas et~al.}]{jiang2023mistral}
Albert~Q. Jiang, Alexandre Sablayrolles, Arthur Mensch, Chris Bamford, Devendra~Singh Chaplot, Diego de~las Casas, and 1 others. 2023.
\newblock Mistral 7{B}.
\newblock \emph{arXiv preprint arXiv:2310.06825}.

\bibitem[{Kipnis et~al.(2025)Kipnis, Voudouris, Schulze~Buschoff et~al.}]{kipnis2025metabench}
Alex Kipnis, Konstantinos Voudouris, Luca~M. Schulze~Buschoff, and 1 others. 2025.
\newblock metabench -- a sparse benchmark of reasoning and knowledge in large language models.
\newblock In \emph{Proceedings of the 13th International Conference on Learning Representations}.

\bibitem[{Perlitz et~al.(2024)Perlitz, Bandel, Gera, Arviv, Shlain, Shmueli-Scheuer, Choshen, Slonim, and Sheinwald}]{perlitz2024efficient}
Yotam Perlitz, Elron Bandel, Ariel Gera, Ofir Arviv, Michal Shlain, Michal Shmueli-Scheuer, Leshem Choshen, Noam Slonim, and Dafna Sheinwald. 2024.
\newblock Efficient benchmarking (of language models).
\newblock In \emph{Proceedings of the 2024 Conference of the North American Chapter of the Association for Computational Linguistics}.

\bibitem[{Polo et~al.(2024)Polo, Weber, Choshen, Sun, Xu, and Yurochkin}]{polo2024tinybenchmarks}
Felipe~Maia Polo, Lucas Weber, Leshem Choshen, Yuekai Sun, Gongjun Xu, and Mikhail Yurochkin. 2024.
\newblock tiny{B}enchmarks: evaluating {LLM}s with fewer examples.
\newblock In \emph{Proceedings of the 41st International Conference on Machine Learning}.

\bibitem[{Reddi et~al.(2020)Reddi, Cheng, Kanter, Mattson, Schmuelling, Wu, Anderson, Breeskin, Bughici, Cebo et~al.}]{reddi2020mlperf}
Vijay~Janapa Reddi, Christine Cheng, David Kanter, Peter Mattson, Guenther Schmuelling, Carole-Jean Wu, Brian Anderson, Maximilien Breeskin, Mark Bughici, Ciro Cebo, and 1 others. 2020.
\newblock {MLPerf} inference benchmark.
\newblock \emph{Proceedings of the ACM/IEEE International Symposium on Computer Architecture (ISCA)}.

\bibitem[{Rubinstein and Kroese(2008)}]{rubinstein2008simulation}
Reuven~Y. Rubinstein and Dirk~P. Kroese. 2008.
\newblock \emph{Simulation and the {M}onte {C}arlo Method}, 2nd edition.
\newblock John Wiley \& Sons.

\bibitem[{Wang et~al.(2026)Wang, Wang, Fu, Min, Feng, Guan, Hu, He, Wang, Yang, Ren, Huang, Liu, and Zhang}]{wang2026essencebench}
Shaobo Wang, Cong Wang, Wenjie Fu, Yue Min, Mingquan Feng, Isabel Guan, Xuming Hu, Conghui He, Cunxiang Wang, Kexin Yang, Xingzhang Ren, Fei Huang, Dayiheng Liu, and Linfeng Zhang. 2026.
\newblock Rethinking {LLM} evaluation: Can we evaluate {LLM}s with 200$\times$ less data?
\newblock In \emph{Proceedings of the 14th International Conference on Learning Representations}.

\bibitem[{Wang et~al.(2024)Wang, Ma, Zhang, Ni, Chandra, Guo, Ren, Arulraj, He, Jiang et~al.}]{wang2024mmlu}
Yubo Wang, Xueguang Ma, Ge~Zhang, Yuansheng Ni, Abhranil Chandra, Shiguang Guo, Weiming Ren, Aaran Arulraj, Xuan He, Ziyan Jiang, and 1 others. 2024.
\newblock {MMLU}-pro: A more robust and challenging multi-task language understanding benchmark.
\newblock \emph{arXiv preprint arXiv:2406.01574}.

\bibitem[{Yang et~al.(2025)Yang, Yang, Zhang et~al.}]{yang2025qwen3}
An~Yang, Baosong Yang, Beichen Zhang, and 1 others. 2025.
\newblock Qwen3 technical report.
\newblock \emph{arXiv preprint arXiv:2505.09388}.

\bibitem[{Yuan et~al.(2025)Yuan, Zhang, Feng, Li, Wang, Shi, Tan, Pan, Hu, and Li}]{yuan2025tailoredbench}
Peiwen Yuan, Yueqi Zhang, Shaoxiong Feng, Yiwei Li, Xinglin Wang, Jiayi Shi, Chuyi Tan, Boyuan Pan, Yao Hu, and Kan Li. 2025.
\newblock Beyond one-size-fits-all: Tailored benchmarks for efficient evaluation.
\newblock In \emph{Proceedings of the 63rd Annual Meeting of the Association for Computational Linguistics}.

\bibitem[{Zhang et~al.(2026)Zhang, Guo, Lu, Dai, Xia, and Wang}]{zhang2026sparseeval}
Taolin Zhang, Hang Guo, Wang Lu, Tao Dai, Shu-Tao Xia, and Jindong Wang. 2026.
\newblock Sparse{E}val: Efficient evaluation of large language models by sparse optimization.
\newblock In \emph{Proceedings of the 14th International Conference on Learning Representations}.

\bibitem[{Zhang et~al.(2025)Zhang, Long, Yang et~al.}]{zhang2025qwen3embedding}
Yanzhao Zhang, Dingkun Long, Yueting Yang, and 1 others. 2025.
\newblock Qwen3 embedding: Advancing text embedding and reranking through foundation models.
\newblock \emph{arXiv preprint arXiv:2506.05176}.

\bibitem[{Zhou et~al.(2024)Zhou, Lu, Mishra, Brahma, Basu, Luan, Zhou, and Hou}]{zhou2023ifeval}
Jeffrey Zhou, Tianjian Lu, Swaroop Mishra, Siddhartha Brahma, Sujoy Basu, Yi~Luan, Denny Zhou, and Le~Hou. 2024.
\newblock Instruction-following evaluation for large language models.
\newblock In \emph{Proceedings of the 62nd Annual Meeting of the Association for Computational Linguistics}.

\end{thebibliography}

\clearpage
\appendix
\section*{Appendix}

\section{Leave-$K$-Out: Full Results}
\label{app:leave_k_out}

This appendix provides the full leave-$K$-out results summarized in Section~\ref{sec:ablation_calibration}.

For MMLU and GSM8K, $n^{*}$ is perfectly stable across all 98 leave-$K$-out combinations (all values of $K$), and the max held-out drift equals that of the single worst-drifting model (1.20\,pp for MMLU, 2.62\,pp for GSM8K).

For IFEVAL, $n^{*}{=}250$ is stable in 87 of 98 combinations. The 11 unstable cases all involve simultaneously removing both Qwen2.5-3B and Mistral-7B from the calibration pool. Table~\ref{tab:lko_ifeval_unstable} lists these combinations.

\begin{table*}[h]
\centering
\small
\begin{tabular}{l c}
\toprule
\textbf{Models held out} & $n^{*}$ \\
\midrule
Qwen2.5-3B, Mistral-7B & 300 \\
Qwen2.5-3B, Mistral-7B, Llama-3.1-8B-Base & 300 \\
Qwen2.5-3B, Mistral-7B, Llama-3.1-8B-Inst & 300 \\
Qwen2.5-3B, Mistral-7B, Qwen3.5-35B-A3B & 300 \\
Qwen2.5-3B, Mistral-7B, Llama-3.1-8B-Base, Llama-3.1-8B-Inst & 300 \\
Qwen2.5-3B, Mistral-7B, Llama-3.1-8B-Base, Qwen3.5-35B-A3B & 300 \\
Qwen2.5-3B, Mistral-7B, Llama-3.1-8B-Inst, Qwen3.5-35B-A3B & 300 \\
Qwen2.5-3B, Mistral-7B, Phi-4-mini & 200 \\
Qwen2.5-3B, Mistral-7B, Phi-4-mini, Llama-3.1-8B-Base & 200 \\
Qwen2.5-3B, Mistral-7B, Phi-4-mini, Llama-3.1-8B-Inst & 200 \\
Qwen2.5-3B, Mistral-7B, Phi-4-mini, Qwen3.5-35B-A3B & 200 \\
\bottomrule
\end{tabular}
\caption{IFEVAL unstable leave-$K$-out combinations where $n^{*}$ differs from the 7-model baseline of 250. All involve removing both Qwen2.5-3B and Mistral-7B; max drift is 5.94\,pp (driven by Mistral-7B). When Phi-4-mini remains in the calibration pool $n^{*}$ rises to 300; when Phi-4-mini is also held out, $n^{*}$ drops to 200.}
\label{tab:lko_ifeval_unstable}
\end{table*}

\section{K-Means Clustering Details}
\label{app:kmeans}

For the k-means clustering strategy in Section~\ref{sec:ablation_strategy}, we embed each benchmark item using Qwen3-Embedding-8B~\citep{zhang2025qwen3embedding} which reports strong performance on MTEB embedding tasks. We encode the full item text (question for MMLU/GSM8K; instruction prompt for IFEVAL) and select $k$ by grid search over candidate values, choosing the $k$ with the highest silhouette score.

\begin{table}[h]
\centering
\small
\begin{tabular}{l c c l}
\toprule
\textbf{Benchmark} & \textbf{$k$ range} & \textbf{Best $k$} & \textbf{Silhouette} \\
\midrule
IFEVAL & 5--30 & 25 & 0.019 \\
MMLU & 5--60 & 55 & 0.048 \\
GSM8K & 5--30 & 25 & 0.028 \\
\bottomrule
\end{tabular}
\caption{Number of clusters $k$ selected by silhouette score for each benchmark.}
\label{tab:kmeans}
\end{table}

The low silhouette scores (0.01--0.05) show that benchmark items do not form tight semantic clusters. This is consistent with our finding that k-means clustering does not improve over random sampling at $n^{*}$ (see \ref{sec:ablation_strategy}). This is because the subset is already large enough for low sampling variance, and the strata structure adds negligible benefit.

\section{IFEVAL Per-Metric Accuracy Breakdown}
\label{app:permodel}

Table~\ref{tab:table1} reports IFEVAL drift on inst\_strict only. Table~\ref{tab:ifeval_allmetrics} shows drift on all 4 metrics for completeness.
All four metrics are consistent: mean $|\delta|$ ranges 0.79--1.00\,pp and max $|\delta|$ ranges 1.52--2.16\,pp, demonstrating that the conclusions from the main table hold across all IFEVAL metrics.

\begin{table}[t!]
\centering
\small
\setlength{\tabcolsep}{2pt}
\resizebox{\columnwidth}{!}{%
\begin{tabular}{l rr rr rr rr}
\toprule
& \multicolumn{2}{c}{\textbf{p\_strict}} & \multicolumn{2}{c}{\textbf{p\_loose}} & \multicolumn{2}{c}{\textbf{i\_strict}} & \multicolumn{2}{c}{\textbf{i\_loose}} \\
\cmidrule(lr){2-3} \cmidrule(lr){4-5} \cmidrule(lr){6-7} \cmidrule(lr){8-9}
\textbf{Model} & Sub & $\delta$ & Sub & $\delta$ & Sub & $\delta$ & Sub & $\delta$ \\
\midrule
Llama-8B-Base & 24.00 & $-$0.21 & 26.40 & +0.71 & 34.03 & $-$1.10 & 35.86 & $-$0.95 \\
Qwen2.5-3B & 61.60 & +0.97 & 65.20 & +0.87 & 70.16 & +0.85 & 73.82 & +0.92 \\
Mistral-7B & 46.80 & $-$1.07 & 53.20 & $-$1.14 & 57.07 & $-$0.85 & 63.35 & $-$0.56 \\
Phi-4-mini & 70.00 & +2.16 & 72.80 & +1.45 & 77.49 & +1.83 & 80.10 & +0.73 \\
Llama-8B-Inst & 73.20 & $-$0.37 & 80.80 & +1.69 & 80.63 & $-$0.31 & 86.65 & +1.52 \\
Qwen3-8B & 82.00 & $-$1.55 & 86.80 & $-$0.08 & 87.43 & $-$1.41 & 90.58 & $-$0.43 \\
Qwen3.5-35B & 84.40 & +0.67 & 86.80 & $-$0.45 & 89.53 & +0.56 & 91.10 & $-$0.39 \\
\midrule
Mean $|\delta|$ & & 1.00 & & 0.91 & & 0.99 & & 0.79 \\
Max $|\delta|$ & & 2.16 & & 1.69 & & 1.83 & & 1.52 \\
\bottomrule
\end{tabular}%
}
\caption{Full per-metric breakdown of IFEVAL subset accuracy (\%) and drift $\delta$ in percentage points, using seed\,42 at $n^{*}{=}250$ with random sampling. Metrics are reported at both prompt level (p\_strict, p\_loose) and instruction level (i\_strict, i\_loose). All four metrics are consistent, with mean $|\delta|$ ranging from 0.79 to 1.00\,pp and max $|\delta|$ from 1.52 to 2.16\,pp.}
\label{tab:ifeval_allmetrics}
\end{table}

\section{Threshold Sensitivity Analysis}
\label{app:tau_sensitivity}

\begin{figure}[h]
    \centering
    \includegraphics[width=0.95\columnwidth]{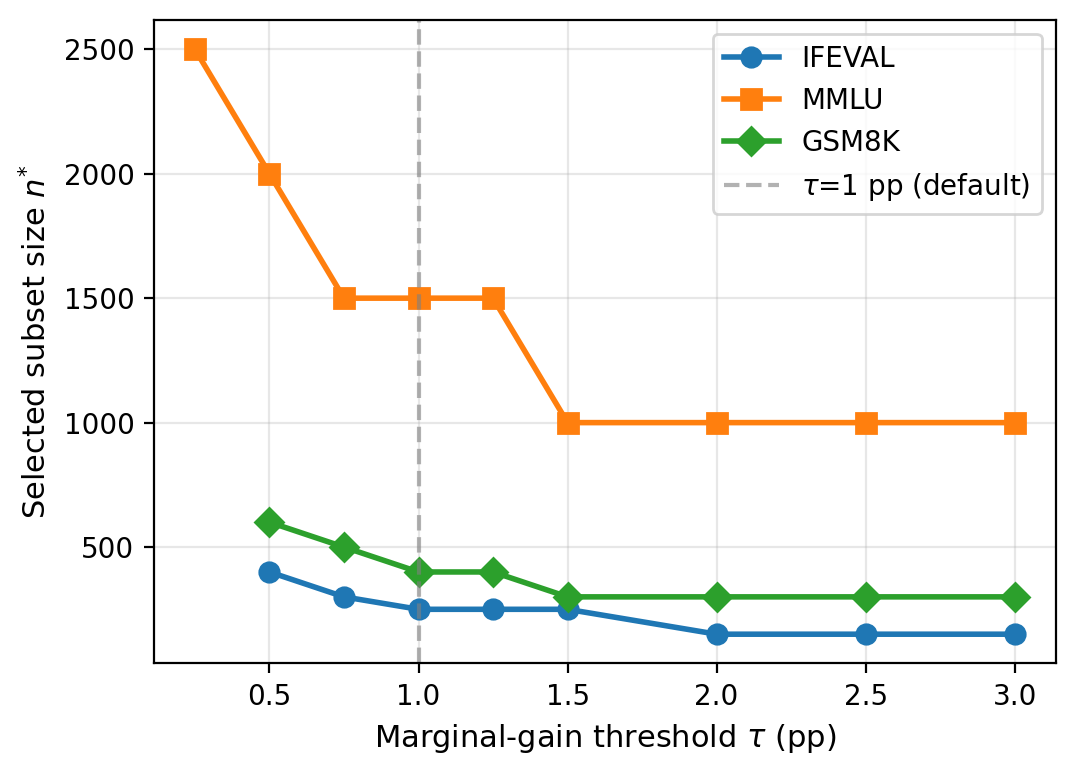}
    \caption{Selected subset size $n^{*}$ as a function of the marginal-gain threshold $\tau$ for each benchmark. Flat segments indicate $\tau$ ranges over which $n^{*}$ is unchanged.}
    \label{fig:tau_sensitivity}
\end{figure}

MINCE selects $n^{*}$ as the first candidate size where the marginal gain in worst-case P95 drift drops below $\tau$. We set $\tau{=}1$\,pp throughout the paper. Table~\ref{tab:tau_sensitivity} sweeps $\tau$ from 0.5 to 2.0\,pp and reports the resulting $n^{*}$, benchmark reduction, and worst-case P95 drift for each benchmark. Figure~\ref{fig:tau_sensitivity} visualizes the relationship between $\tau$ and $n^{*}$.

\begin{table}[h]
\centering
\small
\setlength{\tabcolsep}{3pt}
\begin{tabular}{r | ccc | ccc | ccc}
\toprule
& \multicolumn{3}{c|}{\textbf{IFEVAL} (541)} & \multicolumn{3}{c|}{\textbf{MMLU} (14042)} & \multicolumn{3}{c}{\textbf{GSM8K} (1319)} \\
$\tau$ (pp) & $n^{*}$ & \%$\downarrow$ & P95 & $n^{*}$ & \%$\downarrow$ & P95 & $n^{*}$ & \%$\downarrow$ & P95 \\
\midrule
0.50 & 400 & 26\% & 2.6 & 2000 & 86\% & 2.0 & 600 & 55\% & 3.0 \\
0.75 & 300 & 45\% & 3.8 & 1500 & 89\% & 2.3 & 500 & 62\% & 3.4 \\
\textbf{1.00} & \textbf{250} & \textbf{54\%} & \textbf{4.5} & \textbf{1500} & \textbf{89\%} & \textbf{2.3} & \textbf{400} & \textbf{70\%} & \textbf{4.1} \\
1.25 & 250 & 54\% & 4.5 & 1500 & 89\% & 2.3 & 400 & 70\% & 4.1 \\
1.50 & 250 & 54\% & 4.5 & 1000 & 93\% & 2.9 & 300 & 77\% & 5.0 \\
2.00 & 150 & 72\% & 7.0 & 1000 & 93\% & 2.9 & 300 & 77\% & 5.0 \\
\bottomrule
\end{tabular}
\caption{Sensitivity of $n^{*}$ to the marginal-gain threshold $\tau$, reporting the resulting subset size, benchmark reduction, and worst-case 95th-percentile drift (P95) in percentage points across all models and metrics. The default $\tau{=}1$\,pp used throughout the paper is shown in \textbf{bold}.}
\label{tab:tau_sensitivity}
\end{table}

Three observations support the choice of $\tau{=}1$\,pp.
First, as shown in Table~\ref{tab:tau_sensitivity}, MMLU produces \textit{identical} $n^{*}{=}1500$ across $\tau \in [0.75, 1.25]$\,pp, and GSM8K produces \textit{identical} $n^{*}{=}400$ across $\tau \in [1.0, 1.25]$\,pp, because the marginal gain drops sharply between consecutive candidate sizes with no intermediate values near 1\,pp.
Second, Figure~\ref{fig:tau_sensitivity} shows that IFEVAL's $n^{*}$ is stable across $\tau \in [1.0, 1.5]$\,pp at $n^{*}{=}250$, and varies gradually outside this range---lowering $\tau$ to 0.75\,pp increases $n^{*}$ by only 50 items (to 300), while raising $\tau$ to 2.0\,pp decreases $n^{*}$ to 150.
Third, the P95 drift values in Table~\ref{tab:tau_sensitivity} are conservative upper bounds and 95\% of random subsets at $n^{*}$ produce worst-case drift \textit{below} these values. The 100-seed random-subset evaluation in Figure~\ref{fig:strategy} confirms this: mean $|\delta|$ is ${\sim}$1.5\,pp for IFEVAL---well below the 4.5\,pp P95 bound.
Practitioners can run a similar sweep over $\tau$ values to select a threshold appropriate for their application.

\section{GPU Evaluation Speedup Breakdown}
\label{app:timing}

Table~\ref{tab:timing} reports per-model GPU speedup ratios (full$\div$subset wall-clock time at batch size\,1) for all 7 BF16 calibration models. NPU speedups are reported in Table~\ref{tab:npu} in the main text.
MMLU sees the largest speedups (median 8.1$\times$), consistent with its 89\% size reduction. GSM8K speedups are uniform across models at ${\sim}$3.3$\times$, matching its 70\% reduction. IFEVAL speedups vary more widely (2.1--9.2$\times$) because generation length differs across models.

\begin{table}[h]
\centering
\small
\setlength{\tabcolsep}{4pt}
\begin{tabular}{l ccc}
\toprule
& \multicolumn{3}{c}{\textbf{Evaluation Speedup ($\times$)}} \\
\cmidrule(lr){2-4}
\textbf{Model} & IFEVAL & MMLU & GSM8K \\
\midrule
Llama-3.1-8B (base) & 4.1$\times$ & 8.1$\times$ & 3.2$\times$ \\
Llama-3.1-8B-Inst & 2.7$\times$ & 8.2$\times$ & 3.2$\times$ \\
Phi-4-mini-Inst & 2.6$\times$ & 6.7$\times$ & 3.2$\times$ \\
Mistral-7B-Inst & 2.1$\times$ & 10.5$\times$ & 3.3$\times$ \\
Qwen2.5-3B-Inst & 2.2$\times$ & 5.8$\times$ & 3.3$\times$ \\
Qwen3-8B & 5.8$\times$ & 7.8$\times$ & 3.3$\times$ \\
Qwen3.5-35B-A3B & 9.2$\times$ & 8.9$\times$ & 3.3$\times$ \\
\midrule
Median & 2.7$\times$ & 8.1$\times$ & 3.3$\times$ \\
\bottomrule
\end{tabular}
\caption{GPU evaluation speedup ratios (full$\div$subset wall-clock time, batch size\,1) for 7 BF16 calibration models.}
\label{tab:timing}
\end{table}

\end{document}